\documentclass[conference]{IEEEtran}
\IEEEoverridecommandlockouts
\usepackage{cite}
\usepackage{amsmath,amssymb,amsfonts}
\usepackage{algorithmic}
\usepackage{graphicx}
\usepackage{textcomp}
\usepackage{xcolor}

\def\BibTeX{{\rm B\kern-.05em{\sc i\kern-.025em b}\kern-.08em
 T\kern-.1667em\lower.7ex\hbox{E}\kern-.125emX}}
\begin{document}

\title{Twitter Data Analysis: Izmir Earthquake Case
}

\author{\IEEEauthorblockN{1\textsuperscript{st} Özgür AĞRALİ}
\IEEEauthorblockA{\textit{Department of Artificial Intelligence} \\
\textit{Muğla Sıtkı Koçman University}\\
Muğla, Türkiye \\
agrali.ozgurr@gmail.com}

\and
\IEEEauthorblockN{2\textsuperscript{nd} Hakan SÖKÜN}
\IEEEauthorblockA{\textit{Department of Artificial Intelligence
} \\
\textit{Muğla Sıtkı Koçman University}\\
Muğla, Türkiye \\
hakansokun@posta.mu.edu.tr}
\and
\IEEEauthorblockN{3\textsuperscript{rd} Enis KARAARSLAN}
\IEEEauthorblockA{\textit{Department of Artificial Intelligence} \\
\textit{Muğla Sıtkı Koçman University}\\
Muğla, Türkiye \\
enis.karaarslan@mu.edu.tr}

}


\maketitle

\begin{abstract}
Türkiye is located on a fault line; earthquakes often occur on a large and small scale. There is a need for effective solutions for gathering current information during disasters. We can use social media to get insight into public opinion. This insight can be used in public relations and disaster management. In this study, Twitter posts on Izmir Earthquake that took place on October 2020 are analyzed. We question if this analysis can be used to make social inferences on time. Data mining and natural language processing (NLP) methods are used for this analysis. NLP is used for sentiment analysis and topic modelling. The latent Dirichlet Allocation (LDA) algorithm is used for topic modelling. We used the Bidirectional Encoder Representations from Transformers (BERT) model working with Transformers architecture for sentiment analysis. It is shown that the users shared their goodwill wishes and aimed to contribute to the initiated aid activities after the earthquake. The users desired to make their voices heard by competent institutions and organizations. The proposed methods work effectively. Future studies are also discussed.
\end{abstract}

\section{Introduction}
\begin{figure}[hbt!]
\centering
 \includegraphics[width=.99\linewidth]{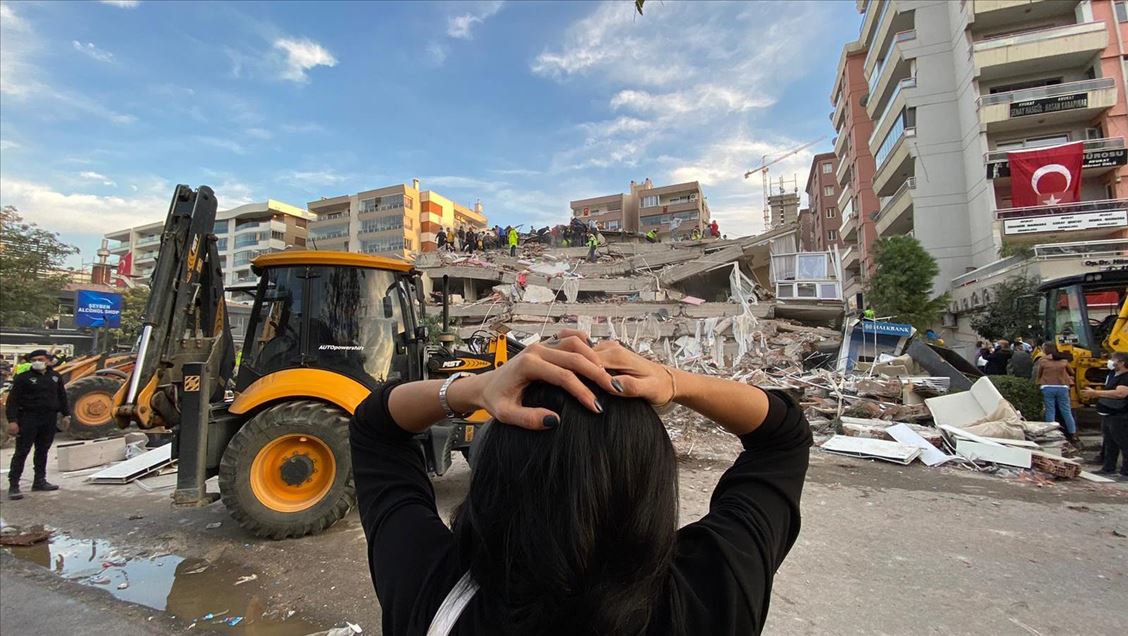}
\caption { \label{fig:deprem-photo}An image obtained on the first day of the earthquake\cite{b21}}
\end{figure}

An earthquake occurred with its epicenter off the Aegean Sea on October the 30th 2020, around 14:51, and the magnitude of the earthquake was 6.9 \cite{b1}. After the earthquake, aftershocks continued in the region. The Izmir earthquake had an impact on many people, both local and in general immediately after it occurred. Great destructions occurred and many people lost their lives. The impact was huge as it is seen in Figure~\ref{fig:deprem-photo}.

The public shared their views, especially through social media platforms. Information about people who have relatives in the region or those who may be under the rubble was also shared quite frequently. For this reason, the shares made on Twitter have gained a kind of notification feature. It has also been a guide for those who carry out search and rescue activities.

It is very difficult to directly process and understand the data received from social media platforms that contain different types of data such as Twitter. Social media data is complex and includes a wide variety of data. Making more detailed data analysis on large datasets and using artificial intelligence technologies in decision making for the future can offer effective solutions. The usage can yield effective results. In order to analyze the obtained data correctly, some operations should be applied on the data. These processes are data manipulation, text pre-processing, and feature extraction. In addition, machine learning and artificial intelligence algorithms can be applied depending on the originality of the dataset. In this way, different estimations and analyzes can be performed. In addition, two of the most widely used classification methods in NLP are used: LDA for topic modeling and BERT model for sentiment analysis. 

This study includes an exploratory analysis of the posts about the Izmir earthquake. In this context, total of 626,384 tweets were collected. The following operations were performed on these collected tweets: singularization, text pre-processing, feature extraction, sentiment analysis, topic modeling, and visualization. Analyzes were made with the help of these processes.

\section{Related Works}

Recently, the processing and analysis of social media data is a topic of great interest. Twitter is one of the most studied social media platforms. E.g; For the analysis of the 2016 US elections \cite{b2} and for the air pollution of the 3 big cities of the USA in 2019 \cite{b3} Twitter data were studied. In these studies, inferences were obtained by making, machine learning methods and visualizations for analysis.

In this study, about the disasters that occurred in the past years \cite{b4}, it has been revealed that using the geographical location in the social media posts can be used to identify the people affected by the disaster. In a recent study \cite{b5}; It was emphasized that social media sharing during the disaster would be important. In addition, it was stated that aid organizations and support teams could contribute in the process.

In various studies \cite{b6,b7,b8,b9,b10,b11}, sentiment analysis was carried out on social media posts and it was aimed to determine people's attitudes towards events. In the current study in which sentiment analysis was performed in Turkish, classical statistical machine learning methods were used \cite{b12}. In another recent study, English sentiment analysis was conducted with the word and rule-based Afinn model, as well as unsupervised learning models such as Textblob and Vader \cite{b13}. In these studies, the analysis was carried out with more grouping and classification processes on the posts.

In our study, it is aimed to analyze the data of the posts made during the disaster and to determine the social impact of the inferences obtained from this analysis. It is aimed to reveal the inferences by evaluating the aims, thoughts and starting points of the post owners.

\section{Materials and Methods}

\begin{figure}[hbt!]
\centering
 \includegraphics[scale=1,width=8.5cm,height=200cm,keepaspectratio]{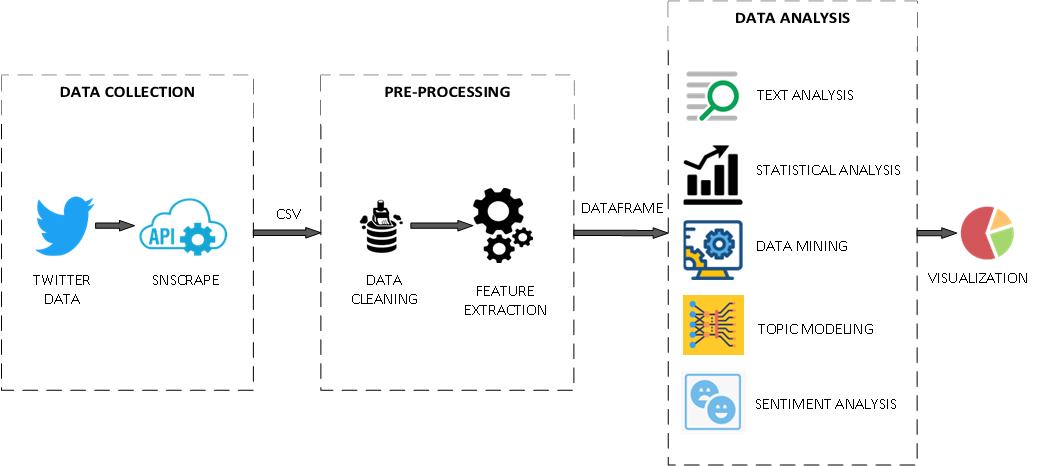}
\caption{ \label{fig:diyagramBig}Data Analysis Flow Diagram}
\end{figure}

The methods used in the study are shown in the workflow diagram in Figure~\ref{fig:diyagramBig}. General methods in the study; Obtaining Twitter data, pre-processing, analyzing and visualizing. These procedures are explained in the relevant subsection.

\subsection{Dataset and Extraction of Data}\label{AA}
There are many web mining methods available to extract data from Twitter. Many free python libraries are available, such as the Twitter API, twitter-scraper, Twint, snscrape and Selenium tools. There are 5 columns on the Twitter dataset. The columns and their descriptions are shown in Table ~\ref{tab:kolonlar}. Also features or attributes that can be pulled from Twitter are detailed in Table~\ref{tab:kolonlar}.
\begin{table}[htbp]
\caption{\label{tab:kolonlar} Dataset Features}
 \centering
 
 \begin{tabular}{| p{0.22\linewidth} | p{0.40\linewidth} | p{0.22\linewidth} | } 
 \hline
 \textbf{Feature Name} & \textbf{Description} & \textbf{Data Type} \\ \hline
 Hashtag & With which tag it was taken & String \\ \hline
Datetime & Posted date and time & Datetime \\ \hline
Tweet Id & Unique value identifying & Float\\ \hline
Text & Content of the post & String \\ \hline
Username & Name of the sharing account & String \\ \hline
 
 \end{tabular}
 \label{tab:my_label}
\end{table}

\subsection{Data Pre-Processing}\label{AA}

The data generated by social media users are unstructured data. This data can also be qualified as “dirty data”. Users may not follow the rules of the language while sharing, they may have expanded their sharing with emoji, or the text may be corrupted due to a software problem. For these and similar reasons, the texts must go through text pre-processing, which is the first step of exploratory data analysis. The pre-processing steps are listed below.

Step 1- Singularization: It is the deletion of duplicate content. The "Tweet Id" field in the dataset will be checked and deleted.

Step 2- Deletion of Unnecessary Data: It is the removal of unnecessary data and personal data from the dataset. No row loss should occur at the end of this process.

Step 3- Deleting the lost data: There must be text in the "Text" attribute, so this is the process of deleting the records where this field is empty.

Step 4- Formatting the date and time field: The "Datetime" field in the dataset is formatted as "YY-MM-DD HH:mm:ss" in accordance with the analysis.

Step 5- Cleaning/editing the text area: The following corrections are made in the text:
\begin{itemize}

\item Deletion of all dirty data types in the share text: link, emoji, special characters,
\item Making all characters lowercase,
\item Cleaning punctuation marks,
\item Removing stop words,
\item Deletion of suffixes of words that have suffixes.
\end{itemize}

\subsection{Feature Extraction}

Exploratory data analysis should be done to derive inferences about shared tweets. The most important stage of exploratory data analysis is feature extraction. For this reason, the feature extraction approach is applied. Technically, this stage is also called Feature Engineering. At this stage, new fields are created by grouping, aggregation and statistical operations on the data.

\subsection{Data Analysis}
Data analysis is carried out in order to obtain significant results on the collected tweets. In this case, data analysis methods are implemented. First of all, what are the most used words and sentences in the sharing and their semantic evaluation are examined. Then, the information about how often and in what time period the shares are made is examined. Finally, the user accounts that share the most and are tagged the most are compared. In this way, significant inferences based on data will be obtained.

In addition, text analysis should be done to obtain analyzes such as for what purpose the shares are made and which topics are most mentioned. Text analysis is one of the most basic purposes of NLP methods. On the other hand, it will be aimed to obtain inferences from numerical and probabilistic analyzes by examining statistical values. Data mining will be used to explain and make sense of the relationships between text and numerical data. On the other hand, the topic modeling method will be applied to evaluate these shares according to their topic headings. Finally, sentiment analysis will be conducted to show in which emotions these shares are made.

\subsection{Data Visualization}
As a result of the analysis, new numerical and text data types are obtained. Different queries are made on these data. Different types of visualizations are made for each data type. It will be aimed to provide earthquake-related determinations by using these visualizations. Another important part is the tags. Tags are the core value in the creation of the dataset. At this stage; Numerical analysis of tags is important.

\section{Implementation}

In this study the Python programming language was used and Google Colab was chosen for development. Web mining tool Snscrape and various machine learning software libraries were used.

The shares on Twitter for the Izmir earthquake were examined and it was evaluated that these shares were made through some tags. For this reason, the most used tags were collected manually. These tags are; '\#deprem', '\#depremizmir', '\#enkazaltında', '\#egededeprem', '\#egedepremi', '\#enkaz', '\#bayraklı', '\#bayraklıdeprem', '\#enkazaltinda', '\#EnkazIhbarIzmir', '\#gecmisolsunizmir', '\#izmirdepremi', '\#yanındayızimir', '\#gecmisolsunizmirim', '\#izmirdeprem', '\#izmiryanindayiz', '\#İzmirDepreminde', '\#izmirdepremi', '\#izmiryanindayiz', '\#izmirgecmisolsun' has been selected.

The data were extracted with the Snscrape library using the specified tag list. The date ranges determined for the collection of data is between October the 30th 2020 and November the 23rd 2020. The day of the earthquake is November the 30th 2020. The date range from which data will be pulled could have been extended further. However, since this study was carried out only on the Izmir earthquake, it was limited to 25 days in order not to deviate from the context.

\begin{table}[htbp]
\caption{\label{tab:oznitelik} The Features obtained after feature extraction}
 \centering
 
 \begin{tabular}{| p{0.35\linewidth} | p{0.5\linewidth} |} 
 \hline
 \textbf{Feature} & \textbf{Description} \\ \hline

	Mentioned users list & List of people involved in the share.\\ \hline
List and number of tags & List and number of tags used. \\ \hline
Internet address & Web address list/number in content. \\ \hline
Raw text word count & Word count used in raw text. \\ \hline
Text word count & Post pre-processing word count. \\ \hline
Number of singular words & Number of singular used words. \\ \hline
Number of stop words & Number of stop words used. \\ \hline
Word length & Avg. word length in pre-processed text. \\ \hline
Raw text character count & Number of characters in the raw text. \\ \hline
Number of text characters & Number of characters in the pre-processed text. \\ \hline
Difference in word count & Difference in word count between text. \\ \hline
 
 \end{tabular}
 \label{tab:my_label}
\end{table}
In total, 626,384 tweet lines and 5 feature columns were collected. These data were obtained with 20 different tags that were shared within 24 days after the earthquake occurred. The pre-processing steps described in the method section were applied step by step on the obtained data set. As a result of the operations, the number of data set records is 390,500 and the number of columns is 6. At this point, it is considered that there are repetitive data on the available data set.

Attribute extraction was performed to make inferences about the shares made. The extracted attributes and their descriptions are listed in the Table ~\ref{tab:oznitelik}. After the feature extraction steps, the number of features in the dataset was 19. The dataset was also run on Orange, which is an open source data mining tool. Table analyzes were performed using more than 2 features.

When it was examined how often the users preferred the tags in the study, it was determined that they mostly used the "\#deprem" tag. The hashtag "\#deprem", where approximately 250 thousand shares were made, is followed by "\#izmirdepremi" and "\#izmirdeprem". It is considered that the posts made mainly focus on these three tags.

\begin{figure}[hbt!]
\centering
 \includegraphics[scale=1,width=8cm,height=8cm,keepaspectratio]{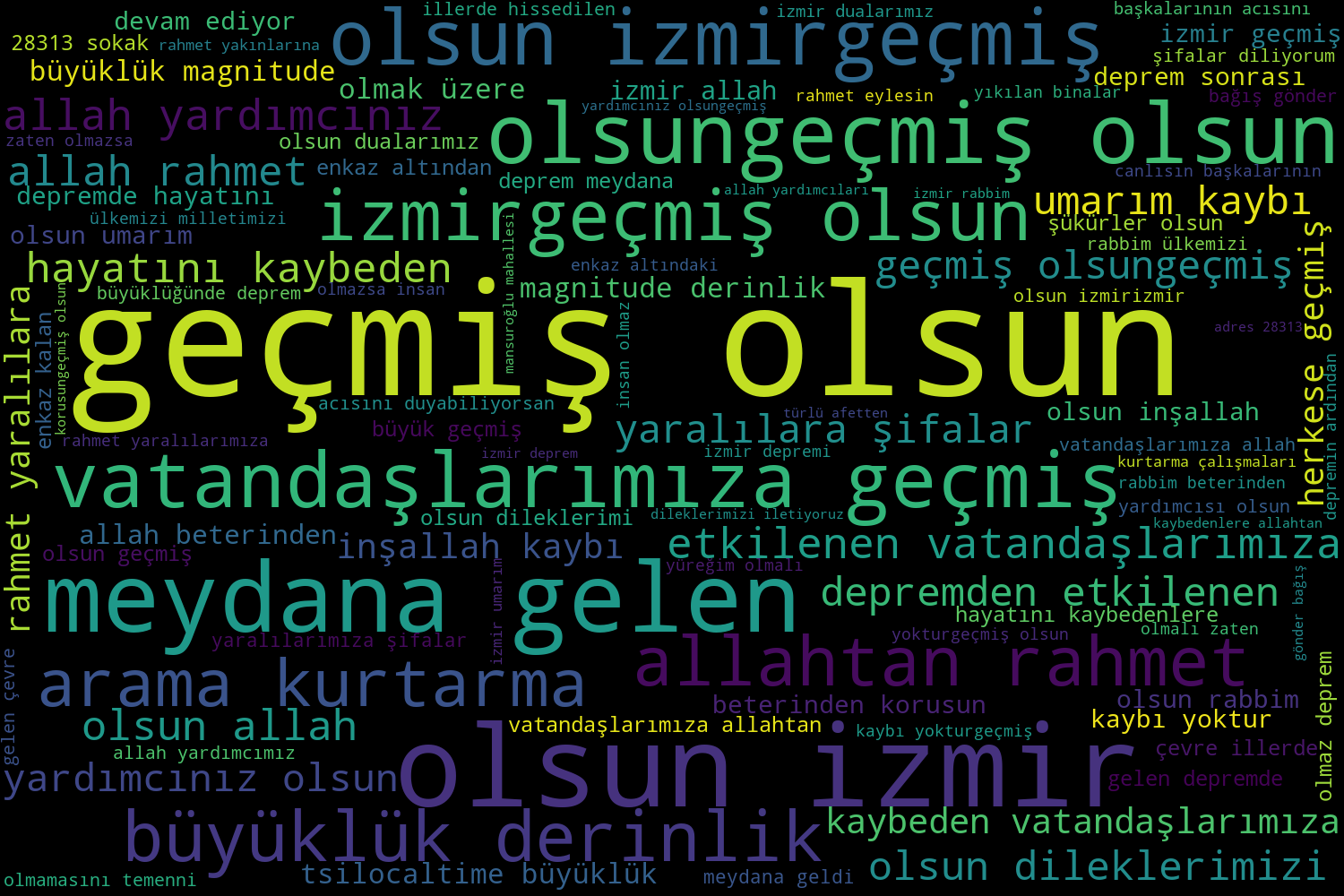}
\caption{ \label{fig:word-cloud} Word cloud of the most common binary words}
\end{figure}

For this reason, an analysis was made about which words were used the most in the data set. It has been observed that the most used singular words are "geçmiş", "olsun", "izmir". The word cloud in the Figure \ref{fig:word-cloud} shows the most commonly used binary words. Accordingly, expressions of wish such as "geçmiş olsun" come first. When the triple word groups containing more understandable sentence expressions are examined, the most used triple word group is "geçmiş olsun izmir". The most used triple word groups, like other word groups, show the atmosphere created by the earthquake. In this case, it can be thought that the requests and wishes of the users dominate the sharing.

After these processes, it is aimed to obtain inferences by making visualizations on the cleaned dataset. In the review of user accounts that shared the most, “TumDepremler” has been the user with more than 2000 shares. The other accounts that share the most are “zelzeleler” and “EMSC”. As a result of the research, it has been understood that these accounts are mostly corporate news sites that give instant earthquake news and accounts that share personal news. It is seen that these users are among the users who tweet the most because they post very frequently and regularly.

In the data set, there are shares starting from the date of the Izmir earthquake on October the 30th 2020, until November the 23rd 2020. 67\% of these tweets were posted in October. This reveals that in the first moments of the earthquake, more information flow has been created by users on Twitter.

When the analysis of the month in which the tweets were posted is analyzed, it has been observed that a very high amount of sharing was made between October the 30th and 31st. In the following days, the interest decreased and the number of tweets decreased day by day. As of November the 3rd, it is seen that much fewer posts have been made. In this context, when the dataset is evaluated as a scope, it is concluded that the most shared time is in the first 5 days.

When examining whether a web address is shared or not in the posts about the Izmir earthquake, 51\% of the posts do not contain a link, while 45\% of them contain 1 web link. This indicates that users did not include any web address in nearly half of the shares.

\begin{figure}[hbt!]
\centering
 \includegraphics[width=.95\linewidth]{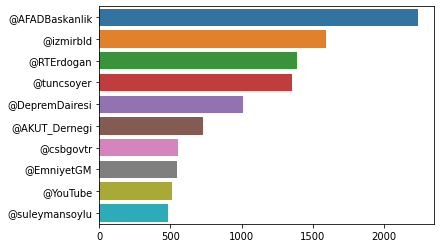}
\caption { \label{fig:mention-liste}Most mentioned usernames}
\end{figure}

In the tweets, the user names mentioned were examined. In this review, it was observed that in most of the shares, users did not tag any users. This explains that users share more to express their feelings.

The most mentioned people in the shared posts are listed in Figure~\ref{fig:mention-liste}. Accordingly, it is seen that users have tagged user accounts that are competent, institutional, and able to contribute to the process, such as "AFADBaskanlik", "izmirbld", and "RTErdogan". When we look at the contents of these posts, it is understood that this was done with the desire to reach the authorities and convey information.

\begin{figure}[hbt!]
\centering
 \includegraphics[width=.99\linewidth]{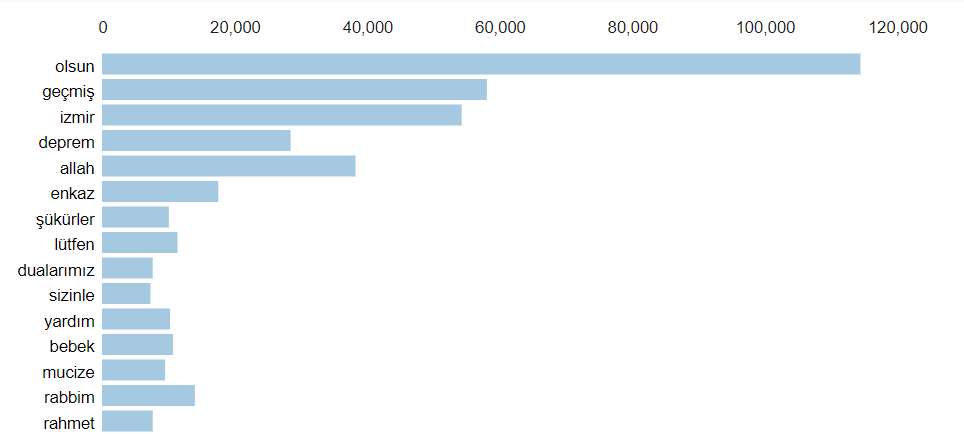}
\caption { \label{fig:topic-modeling}Most mentioned topics}
\end{figure}

As it is known, after cleaning tweet texts (removing; stopwords, emojis, punctuation marks, etc.), a smoother text is obtained. To make a more detailed analysis of the shares, these cleaned texts are topic modeling work. Topic Modeling is one of the most used text classification and analysis methods in NLP. In this way, the main theme of the texts is reached in general.

In this study, LDA model was applied to the text contents for the topic modeling analysis. LDA performs word and document analysis using Bayesian theory. It is an unsupervised classification model that tries to predict which word represents which topic in which document by evaluating the words and documents as a whole and separately. The model output made as words and documents for this study is shown in Figure~\ref{fig:topic-modeling}. Accordingly, a classification was made as 15 titles. Although similar results were obtained with the results in the word cloud, it shows that "yardım" and "bebek" topics are covered a lot.

Finally, sentiment analysis of the shares was made. Sentiment analysis is a classification problem. As it is known, BERT has achieved great success in the field of NLP, especially for text classification. The Turkish NLP community contributed to the Turkish classification study by developing the BERTurk model. BERTurk is a community-driven BERT model for Turkish.

Two models (pre-trained and fine-tuned) are used in this study. Comparative details of these models are given in Table~\ref{tab:sentiment-models}.

\begin{table}[h]
\caption{\label{tab:sentiment-models} Pre-trained and Fine-tuned Model Information} 
\centering 
\begin{tabular}{l c c rrrrrrr} 
\hline\hline 
\\[0.5ex] Model &Class &Epoch &Dataset &Min. Loss &Accuracy
\\ [1.5ex]
\hline 
\\[0.5ex]
\raisebox{1.5ex}{Pre-trained} & \raisebox{1.5ex}{2}&\raisebox{1.5ex}{3}&\raisebox{1.5ex}{48.290}&\raisebox{1.5ex}{0.16}&\raisebox{1.5ex}{95\%}\\[1ex]

\raisebox{1.5ex}{Fine-tuned} & \raisebox{1.5ex}{3}&\raisebox{1.5ex}{3}&\raisebox{1.5ex}{492.782}&\raisebox{1.5ex}{0.25}&\raisebox{1.5ex}{94\%}\\[1ex]

\hline 
\end{tabular}
\label{tab:PPer}
\end{table}

A pre-trained "bert-base-turkish-sentiment-cased" model was used for BERTurk-based sensitivity analysis \cite{b14}. This model is trained with only two classes (negative and positive) data. In this study, a new sentiment analysis classification model that can fine-tune this pre-trained model and make three-class predictions is trained. For the new model obtained, a data set with three class labels (negative, neutral, positive) was used \cite{b15}. This dataset contains content where people openly express their feelings, such as Wikipedia and online product reviews. In this case, 90\% of this data set is used as training data and 10\% as test data in the model. In addition to this, the model trained by Pytorch is set to eval\_step: 50, learning\_rate: 5e-7 and batch\_size: 4. In addition, experiments have observed that hyperparameters affect the success of this model. Accordingly, the fine-tuned model provided a 94\% success rate in the test data. 

The fine-tuned model had almost the same accuracy as the pre-train with only binary classification. Comparative details of the models are given in Table~\ref{tab:sentiment-models}. After several trials, it was observed that these values were optimal.

\begin{figure}[hbt!]
\centering
 \includegraphics[width=.99\linewidth]{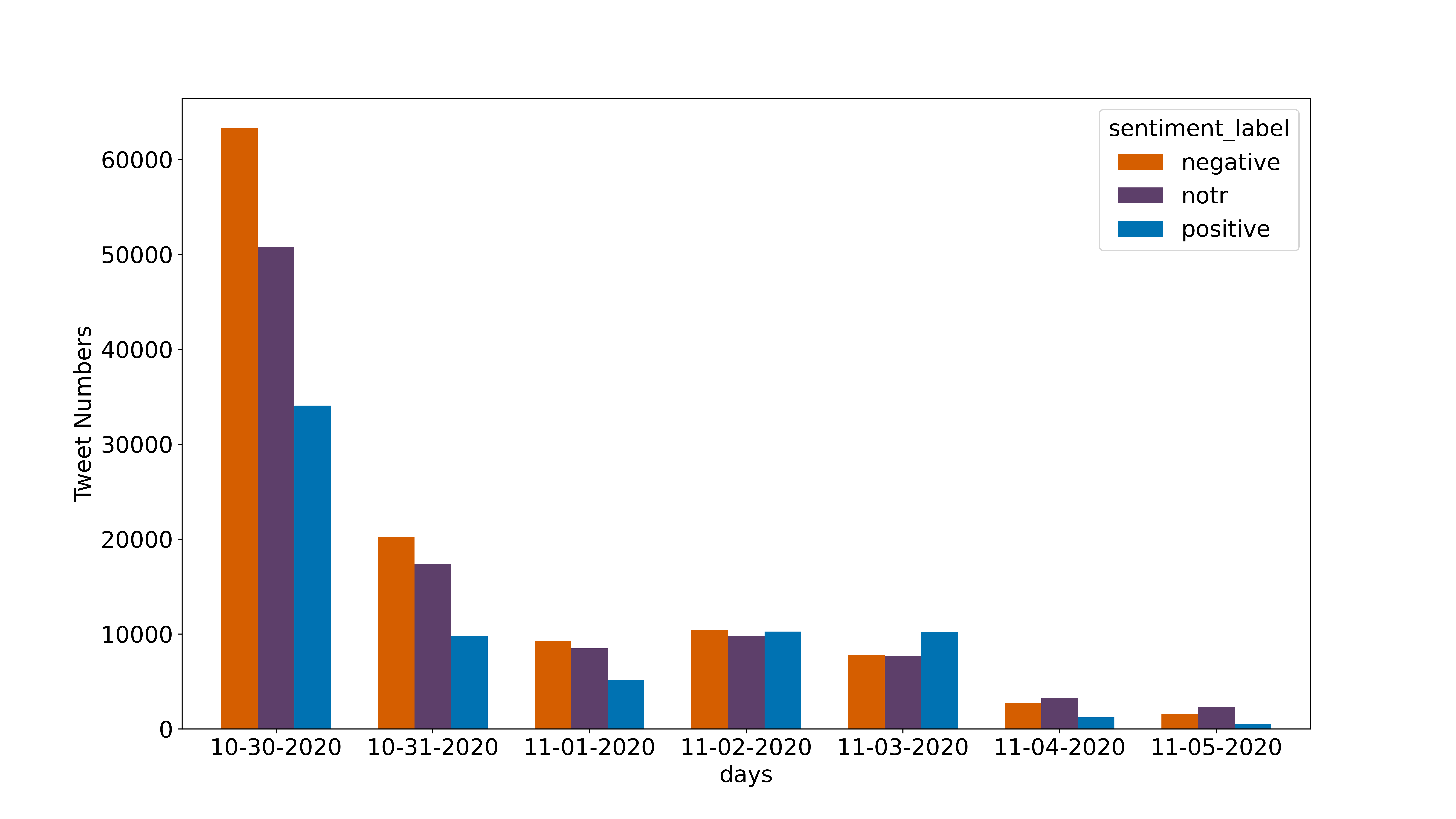}
\caption { \label{fig:sentiment-analysis}Number of tweets and sentiment analysis from day to day}
\end{figure}
Subsequently, cleaned tweet texts are given as input to this model. It provided dual sentiment classification as positive, neutral and negative as output. It can be seen in Figure~\ref{fig:sentiment-analysis} that the great majority of these shares are made negative. In addition, it is seen that the most tweeted days are the first five days after the earthquake.
On the first day of the earthquake, this 8-floor Rıza Bey apartment, which has more than 40 flats and many workplaces in the Bayraklı district of İzmir, was completely destroyed \cite{b16}. For this reason, search and rescue efforts continued for days, especially in this apartment. However, these searches were suspended on November the 1st, 2020. As a result, there was a decrease in the number of tweets posted on November the 1st \cite{b17}. In this case, there was a significant increase in the number of positive tweets, equal to the number of negative tweets. This is because 14-year-old İdil was rescued from the wreckage (58 hours after the earthquake)\cite{b18}.

On November the 3rd (91 hours after the earthquake), 3-year-old baby Ayda was pulled from the rubble \cite{b19}. With this news, there is no decrease in the number of tweets, and it is seen that positive tweets are more than in the first days. For the first time, the number of positive tweets in a day exceeded the number of negative tweets.

Search and rescue efforts ended on November the 4th, 2020 \cite{b20}, and in the following days, the number of tweets gradually decreased, and the number of negative and neutral tweets increased.

\section{Discussion}
Most of the posts about the earthquake were made in the first five days. While people conveyed their wishes in the first days, they shared more news and aid issues in the following days. The classification is made based on the topic modelling results. It shows that the people affected by the earthquake wish well, share aid and follow the earthquake news up-to-date. The results of the sentiment analysis show that the comments mainly express their sadness about the rescued or harmed people. It is understood that everyone expresses their sadness when bad news is shared during the earthquake, and everyone shares their joy when good news is shared. Without any good or bad developments, neutral content tweets were shared.

The reason why more than half of the shares in the data set do not include any web address is that the vast majority of them are not shared for advertising and news purposes. It is observed that users mostly share using the "\#deprem" tag, and the posts are made especially during lunch hours. The users mainly prefer the most straightforward words to describe the event. Mostly good wishes are made.

The users mainly mention corporate accounts in their posts. Such tweets are analyzed, and it is seen that these institutions are in an effort to contribute to the process and primarily respond to the demands.
The most mentioned state officials' accounts were examined, and it is seen that these institutions share very little when compared to aid institutions. This situation shows that aid organizations are in two-way communication while state officials are in one-way communication regarding the requests and aid posts on social media.

\section{ Conclusion}

We showed that social media posts could be analyzed to understand the effects of disasters on society and be used to produce solutions. This preliminary study demonstrates the benefits of using artificial intelligence techniques in this process. Similar systems can be deployed to foresee people's feelings and opinions in another possible disaster. Government officials and aid organizations can use these to make instant inferences. They will be able to respond to people's needs more quickly and systematically. 

This study is about a current and local event. Creative and practical results have emerged as the applied methods and analyzes are very diverse. It is aimed to get more valuable and multi-class data by conducting more joint studies with social sciences to obtain more comprehensive results in future studies. Comprehensive and specific statistical inferences can be made with these data. Pre-trained BERT models, deep learning, and NLP techniques can be used to perform more than three sentiment classifications and more precise analyses for sentiment analysis.

\vspace{12pt}


\begin{thebibliography}{00}

\bibitem{b1} A.C Yalciner, et al. "The 30 October 2020 (11: 51 UTC) Izmir-Samos earthquake and tsunami: post-tsunami field survey preliminary results," in Middle East Technical University, Ankara, Turkey, 2020.


\bibitem{b2} A. Bovet, F. Morone and H. A. Makse, "Validation of Twitter opinion trends with national polling aggregates: Hillary Clinton vs Donald Trump," Scientific reports, 2018, 8.1: 1-16.


\bibitem{b3} S. Gurajala, S Dhaniyala., and J. N. Matthews, "Understanding public response to air quality using tweet analysis," Social Media+ Society, 2019, 5.3.


\bibitem{b4} N. Kankanamge, T. Yigitcanlar, A. Goonetilleke and M. Kamruzzaman, "Determining disaster severity through social media analysis: Testing the methodology with South East Queensland Flood tweets," International journal of disaster risk reduction 2020, 42: 101360.


\bibitem{b5} N. Pourebrahim, S. Sultana, J. Edwards, A. Gochanour, and S. Mohanty, "Understanding communication dynamics on Twitter during natural disasters: A case study of Hurricane Sandy," International journal of disaster risk reduction, 2019, 37: 101176.

\bibitem{b6} A. Ceron, L. Curini, S.M. Iacus and G. Porro, "Every tweet counts? How sentiment analysis of social media can improve our knowledge of citizens," political preferences with an application to Italy and France. New media and society, 2014, 16.2: 340-358.


\bibitem{b7} V. Kharde and P. Sonawane, "Sentiment analysis of twitter data: a survey of techniques," arXiv preprint arXiv:1601.06971, 2016.


\bibitem{b8} A. Shelar, C.Y. Huang, "Sentiment analysis of twitter data," In: 2018 International Conference on Computational Science and Computational Intelligence (CSCI). IEEE, 2018. p. 1301-1302.


\bibitem{b9} M. Albayrak, K.topal and V. Altıntaş, "Sosyal medya üzerinde veri analizi: Twitter," Süleyman Demirel Üniversitesi İktisadi ve İdari Bilimler Fakültesi Dergisi, 2017, 22.Kayfor 15 Özel Sayısı: 1991-1998.


\bibitem{b10} A. Karami, V. Shah, R. Vaezi and A. Bansal, "Twitter speaks: A case of national disaster situational awareness," Journal of Information Science, 2020, 46.3: 313-324.

\bibitem{b11} K. Zahra, M. Imran, and F. O. Ostermann, "Automatic identification of eyewitness messages on twitter during disasters," Information processing \& management, 2020, 57.1: 102107.
\bibitem{b12} F. DALKILIÇ, \& A. Çam, (2021). "Automatic Movie Rating by Using Twitter Sentiment Analysis and Monitoring Tool. Journal of Emerging Computer Technologies," Journal of Emerging Computer Technologies, 2021, 1.2: 55-60.

\bibitem{b13} Ö. AĞRALI, \& Ö. AYDIN, "Tweet Classification and Sentiment Analysis on Metaverse Related Messages," Journal of Metaverse, 2021, 1.1: 25-30.

\bibitem{b14} S. Yıldırım, savasy/bert-base-turkish-sentiment-cased · Hugging Face, 2020. [Online]. Available: https://huggingface.co/savasy/bert-base-turkish-sentiment-cased. [Accessed: 02- Nov- 2020].

\bibitem{b15} B. Ayhan, winvoker/turkish-sentiment-analysis-dataset · Hugging Face, 2022. [Online]. Available: https://huggingface.co/datasets/winvoker/turkish-sentiment-analysis-dataset. [Accessed: Nov- 2022].

\bibitem{b16} “İşte birçok kişiye mezar olan Rıza Bey apartmanı”, 2020. [Online]. Available: https://www.turkiyegazetesi.com.tr/gundem/742966.aspx. [Accessed: Oct. 30, 2020].

\bibitem{b17} T. Albay, 'İzmir'de deprem sonrası yıkılan Rıza Bey Apartmanı enkazındaki kurtarma çalışmasına ara verildi', 2020. [Online]. Available: https://www.aa.com.tr/tr/turkiye/izmirde-deprem-sonrasi-yikilan-riza-bey-apartmani-enkazindaki-kurtarma-calismasina-ara-verildi-/2027207. [Accessed: 01- Nov- 2020].



\bibitem{b18} T. Albay, 'İzmir'deki depremden yaklaşık 58 saat sonra 14 yaşındaki İdil enkazdan yaralı çıkarıldı', 2020. [Online]. Available: https://www.aa.com.tr/tr/turkiye/izmirdeki-depremden-yaklasik-58-saat-sonra-14-yasindaki-idil-enkazdan-yarali-cikarildi/2027463. [Accessed: 02- Nov- 2020].

\bibitem{b19} "Son dakika: 3 yaşındaki Ayda bebek 91'inci saatte enkazdan çıkarıldı", 2020. [Online]. Available: https://www.haberturk.com/son-dakika-haberi-tunc-soyer-acikladi-ekipleri-bir-bebege-canli-ulasti-2857124. [Accessed: 03- Nov- 2020].

\bibitem{b20} "İzmir depremi: Arama kurtarma çalışmaları sona erdi, can kaybı 114'e yükseldi", 2020. [Online]. Available: https://www.bbc.com/turkce/haberler-turkiye-54810440. [Accessed: 04- Nov- 2020].

\bibitem{b21} L. İlhan "İzmir'de deprem", 2020. [Online]. Available: https://www.aa.com.tr/tr/pg/foto-galeri/izmir-de-deprem/0. [Accessed: 30- Oct- 2020].

\end{thebibliography}
\end{document}